\documentclass{article}

     \PassOptionsToPackage{numbers, compress}{natbib}

\usepackage[final]{neurips_2022}

\usepackage[utf8]{inputenc} 
\usepackage[T1]{fontenc}    
\usepackage{hyperref}       
\usepackage{url}            
\usepackage{booktabs}       
\usepackage{amsfonts}       
\usepackage{nicefrac}       
\usepackage{microtype}      
\usepackage{xcolor}         
\usepackage{graphicx}
\title{Protein language model rescue mutations highlight variant effects and structure in clinically relevant genes}

\author{%
  Onuralp Soylemez\textsuperscript{\textdagger} \\
  \texttt{onuralp@gmail.com} \\
  \And
  Pablo Cordero\textsuperscript{\textdagger} \\
  \texttt{pablo@stripe.com} \\
}

\begin{document}

\maketitle

\begin{abstract}
  Despite being self-supervised, protein language models have shown remarkable performance in fundamental biological tasks such as predicting impact of genetic variation on protein structure and function. The effectiveness of these models on diverse set of tasks suggests that they learn meaningful representation of fitness landscape that can be useful for downstream clinical applications. Here, we interrogate the use of these language models in characterizing known pathogenic mutations in curated, medically actionable genes through an exhaustive search of putative compensatory mutations on each variant's genetic background. Systematic analysis of the predicted effects of these compensatory mutations reveal unappreciated structural features of proteins that are missed by other structure predictors like AlphaFold. While deep mutational scan experiments provide an unbiased estimate of the mutational landscape, we encourage the community to generate and curate rescue mutation experiments to inform the design of more sophisticated co-masking strategies and leverage large language models more effectively for downstream clinical prediction tasks. 
\end{abstract}

\section{Introduction}

Understanding the effects of genetic variation in modulating disease is a central task in clinical genetics and genomic medicine, where the ultimate goal is to detect, quantify, and characterize the pathogenicity of particular mutations to elicit correct diagnosis and inform treatments. Recent advances in protein sequence and structure modeling are beginning to show promise to aid in this task. Protein language models have been shown to harbor variant effect information and high accuracy protein structure prediction \citep{alphafold} has vastly expanded the ways protein structure can be used in connecting genetic variation with disease effects. 

More precisely, recent work (\cite{proteinbert}, \cite{transformer_plm}, \cite{eve}) has shown that protein language models can effectively model deep mutational scan data from extensive genotype-phenotype mapping (e.g., fitness landscape of green fluorescent protein \cite{gfp}) and are also capable of predicting pathogenicity of disease-associated mutations (\citep{nadav}, \cite{esm1v}) without further training, simply relying on the underlying patterns mined by self-supervised language modeling. Here, we expand on these trends by interrogating protein language models through \textit{in silico} mutation. 

We leverage an evolutionary insight from compensatory molecular evolution and describe a novel approach to recovering spatial features of protein structures. To ground the approach, we compare compensatory scores with compensated pathogenic deviations (CPDs): pathogenic amino acid substitutions in humans where the human pathogenic state appears to be wild-type in a functionally-equivalent protein from an orthologous species without any drastic fitness impact on the latter genetic background (\cite{dobzhansky}, \cite{cpd}). Additionally, we delve into the patterns of such protein language model rescue mutations and find that they segregate guided by protein structure and can sometimes pinpoint structural features missed by structure predictors like AlphaFold. 

\section{Data and Methods}
\label{methods}

\paragraph{Clinically relevant genetic variants.} We retrieved the latest list of medically actionable genes curated by the American College of Medical Genetics and Genomics (ACMG)\cite{acmg_v3.1}. These genes harbor high penetrance, large effect pathogenic mutations associated with clinically actionable medical conditions. We limited our analysis to 53 genes with less than 1024 amino acid residues in length in line with the ESM-1v pre-training setup \cite{esm1v} (see Appendix for the gene list). For each gene, we extracted ClinVar \cite{clinvar} variants with their corresponding clinical significance annotation, and grouped the variant impact into three categories: pathogenic/likely pathogenic (P/LP), benign/likely benign (B/LB) and variants of unknown significance (VUS). In case of conflicting interpretations of pathogenicity, annotation with the higher number of ClinVar submissions is considered.

For each ClinVar variant, we parsed the global allele frequency for the mutant  pathogenic allele from the gnomAD v2.1 dataset that contains human genetic variation data from 125,748 whole exomes and 15,708 whole genomes \cite{gnomad}. In case of multiple nucleotide changes corresponding to the same protein change, we kept the allele frequency for the more common alternate allele.

\paragraph{Compensated pathogenic deviations (CPDs).} We compiled a list of compensated pathogenic deviations, where the pathogenic ClinVar variant in human is conserved in an ortholog in another species. Presence of such substitutions - even in highly conserved regions - suggest that there must be other amino acid changes either within the same protein or in an interacting protein to mitigate the fitness impact, and these interdependent or epistatic interactions may represent functional or spatial constraints on the corresponding sites. Specifically, using highly conserved multiple sequence alignment from placental mammals, we identified sites where the disease associated ClinVar amino acid state (e.g., His in Arg13His substitution) appears to be wild-type in a functionally equivalent (orthologous) sequence in at least one placental mammal. To assess the relevance of sequence context in the vicinity of the pathogenic mutation, we also identified CPDs where neighboring amino acid residues are completely conserved in the placental mammal phylogeny, likely corresponding to regions of functional importance.

\paragraph{Fitness impact of secondary mutations.} We use the ESM protein language model to score putative secondary mutations on the genetic background of a known disease mutation. Specifically, we interrogate the log odds of each amino acid in each sequence position of the protein language model under the background of the genetic variant of interest (the so-called "wild-type marginal" effect). Independently, such a score highlights the fitness impact of secondary mutations and identifies rescue mutations whose average fitness impact can compensate the fitness reduction caused by the original mutant. In aggregate, summary statistics of such scores in any given position can yield signals of gain or loss of fitness as a result of the background, pathogenic genetic variant and may lead to to insights into structure. Thus, we also consider the z-score across background mutation position of these ESM scores to compare the fitness impact of that background mutation against all others.

\section{Results}
\label{results}

\paragraph{Stability of predictions between different language models.} For any useful clinical application, it is important that pretrained protein language models with different modeling perplexity yield robust predictions. Here we evaluate the consistency of the pathogenicity predictions from two state-of-the-art transformer protein language models. Specifically, we score the pathogenicity of ClinVar variants in \emph{LDLR} gene using  ESM-1v and ESM-2 pre-trained models, and show that there is very strong correlation among the two predictors (Pearson's correlation coefficient $r$=0.91) (see Figure 1). While the correlation attenuates when all ClinVar variants across the entire gene list are considered, two models generate consistent clinical significance annotations (see Supplementary Figure 1).

Notably, variants with unknown significance (VUS) and pathogenic variants (P/LP) show relatively moderate correlation (Pearson's $r$ of 0.60 and 0.59, respectively) when compared to correlation for benign variants (B/LB) between the two language models. Moreover, we found that pre-trained models with more layers scored a subset of VUS in BAG3 more pathogenic. These findings underscore the importance of model selection for \textit{in silico} pathogenicity predictions of clinically relevant variants with unknown significance in medically actionable genes such as \textit{BAG3} (see Supplementary Figure 3).

\begin{figure}
    \centering
    \includegraphics[width=80mm]{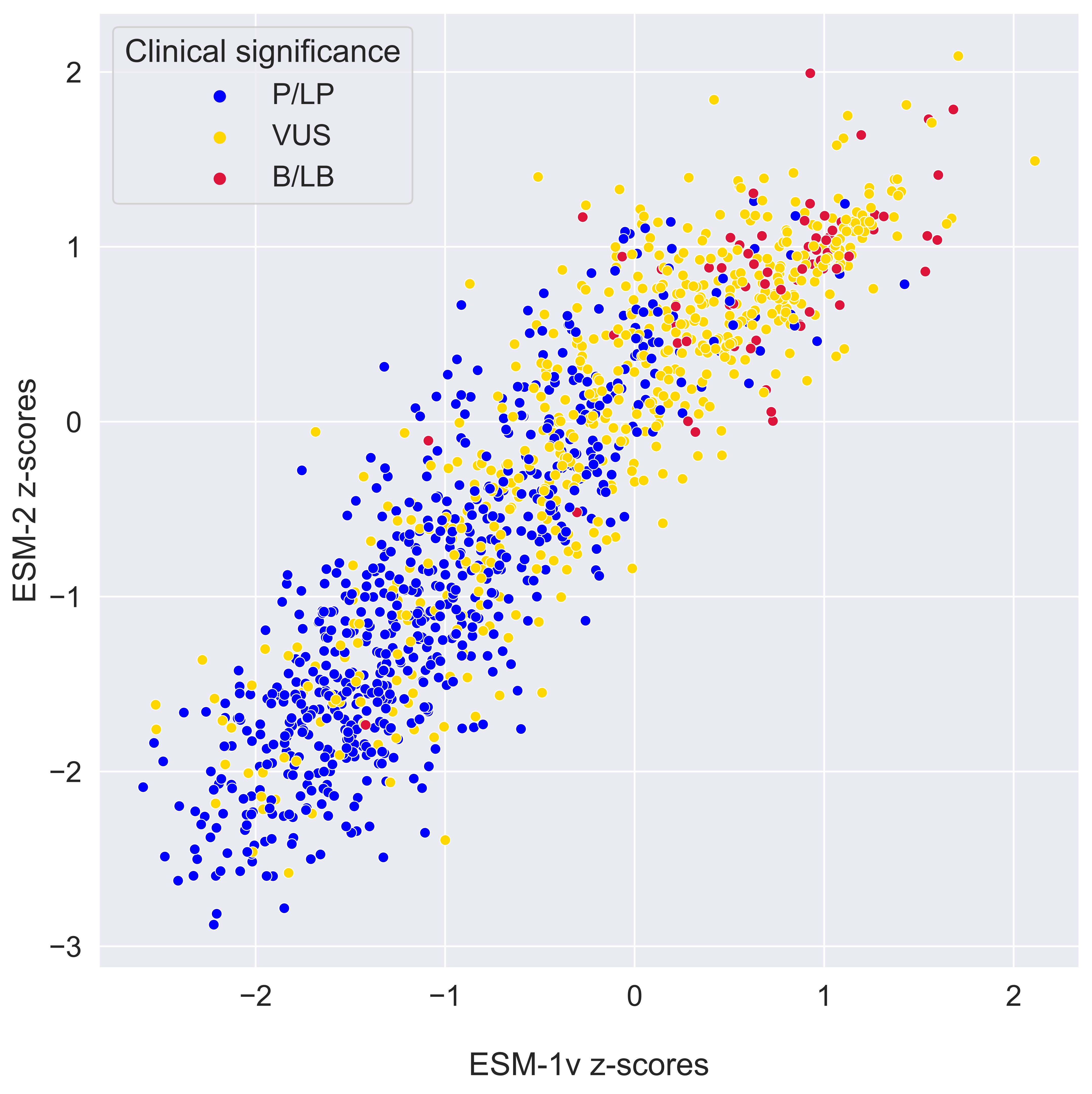}
    \caption{ESM scores are reasonably stable between model versions (Pearson's correlation coefficient ($r$) of 0.91). Here, we compare normalized ESM scores of ESM-1v and ESM-2 for the low-density lipoprotein receptor (\textit{LDLR}) gene which has the highest number of mutations with ClinVar annotations available in ACMG dataset. ESM-1v scores are averaged across five models.}
    \label{fig:esm_score_stability}
\end{figure}

\paragraph{Variant pathogenicity predictions.} Due to natural selection pressure on deleterious alleles that reduce fitness, we expect the predicted pathogenicity scores for the pathogenic missense variants (P/LP) to be inversely correlated with the allele frequency of the mutant alelle. We mapped each ClinVar variant to available large-scale human genetic variation data in gnomAD database, and found that the language model variant impact predictions are consistent with the prevalence of corresponding mutant alleles in the general population (See Figure 2). As expected, predicted scores for the benign missense variants (B/LB) do not show the same correlation. Prediction scores calculated using ESM-2 model show the same trend (see Supplementary Figure 2).

\paragraph{Sequence context around putative rescued sites.} Structural and biochemical analysis of compensated pathogenic mutations (CPDs) found previously that CPDs are on average less deleterious than non-compensated pathogenic mutations \cite{modesto}. To test whether language model predictions recapitulate this empirical observation, we identified CPDs as described in Methods and compared the prediction scores for CPDs against non-CPDs. We found that the distributions of ESM-2 prediction scores of CPDs and non-CPDs are statistically different (Mann-Whitney-Wilcoxon two-sided test p-value < 1e-03), and CPDs appear to be more tolerated than non-CPDs in line with the biochemical analysis (see Supplementary Figure 4). 

Taking into consideration potential misalignment errors in multiple sequence alignments, we repeated the analysis limiting CPDs to sites where the neighboring sites are required to be fully conserved across the phylogeny. Interestingly, we did not observe a statistically significant difference between CPDs and non-CPDs when using such constraint on local homology. This finding may suggest that the existing protein language models may be limited to capture local sequence context when putative compensated mutations are present, and further hint at potential gains from more sophisticated co-masking strategies during pre-training. Modest correlation between ESM-2 model scores and Cross-Protein Transfer (CPT) model \cite{cpt} scores highlight the importance of better understanding the subset of predictions where these models do not agree, and compensatory framework may be helpful to diagnose the inherent limitations of protein language models for resulting in such disagreements (see Supplementary Figure 5).

\begin{figure}
    \centering
    \includegraphics[width=80mm]{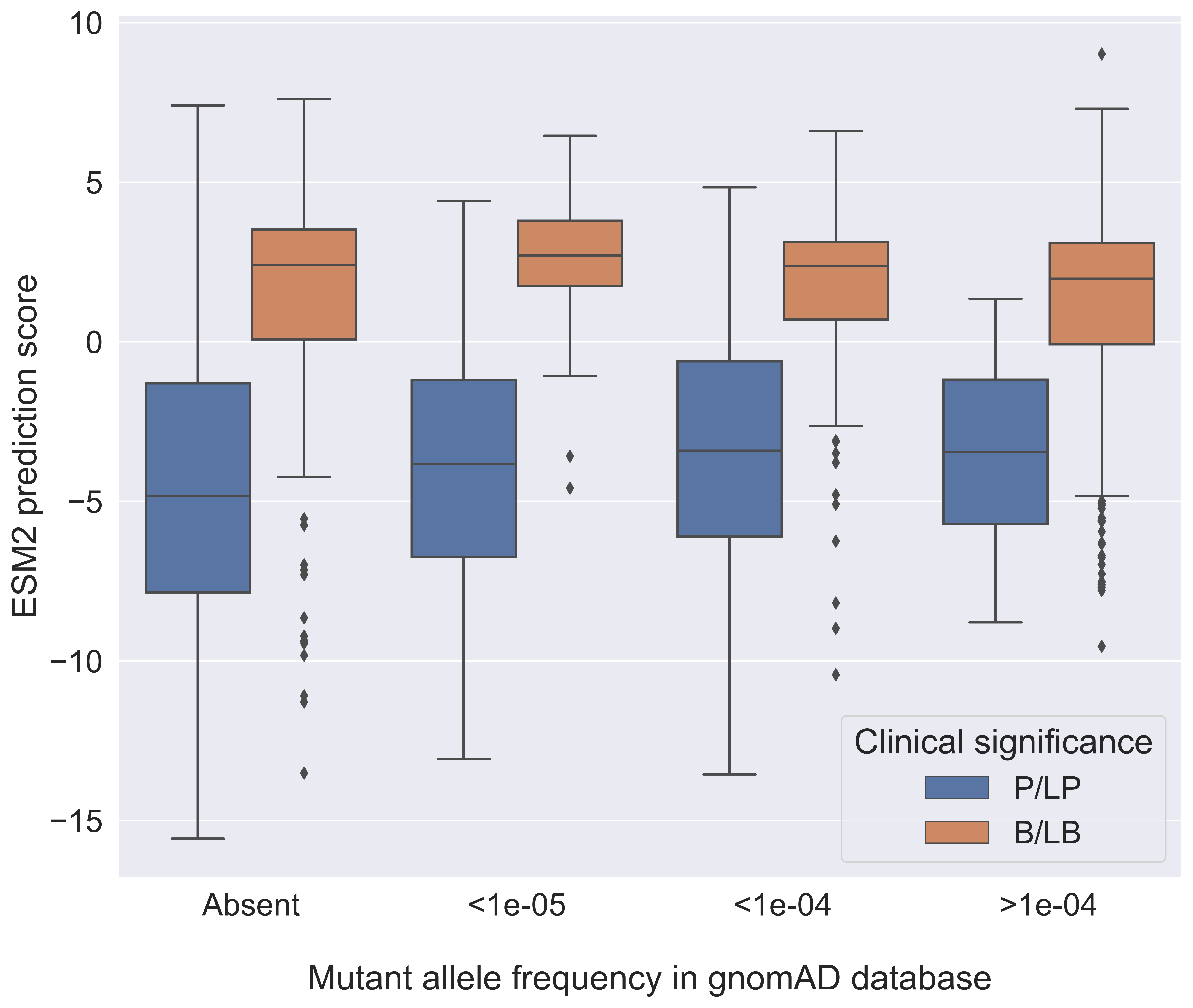}
    \caption{ESM-2 differentiates between pathogenic and benign genetic variants in clinically-relevant genes and this effect grows weaker as the variants are more common in the general population. For all pairwise comparisons, two-sided Mann-Whitney U test p-values < 1e-05.}
    \label{fig:esm_pathogenic_vs_benign}
\end{figure}

\paragraph{Rescue mutation effects reveal unappreciated structural features}. We next interrogated whether simple summary statistics of mutations conditioned against a background variant along the protein sequence revealed any informative compensatory features. This follows the same intuition of residue co-evolution, where compensatory effects arise from sequence perturbations and can crucially inform downstream tasks such as structure prediction. We took the mean wild type marginal score per position, per background variation and z-scored them across all genetic variants. Plotting these scores in aggregate revealed patterns that matched predicted contact maps of the protein's AlphaFold-predicted structures, confirming the functional relevance of these compensatory effects and in line with previous results observing that protein language models can be unsupervised structure predictors. Crucially, in some cases we observed that compensatory effects of these rescue mutations predicted structural features that may have been missed in AlphaFold. For example, the myopathy-related BAG Cochaperone 3 (BAG3) gene is predicted to be mostly unstructured save for one small BAG domain by AlphaFold -- the same domain that has been experimentally characterized. Rescue mutation effects reveal compensatory changes within this supposedly unordered region that harbors multiple variation of unknown significance. Further, comparing ESMfold vs AlphaFold structures confirm that the AlphaFold structure is mostly predicted to be disordered while ESMfold shows a more compact structure with more structural features.

\begin{figure}
  \centering
  \includegraphics[width=135mm]{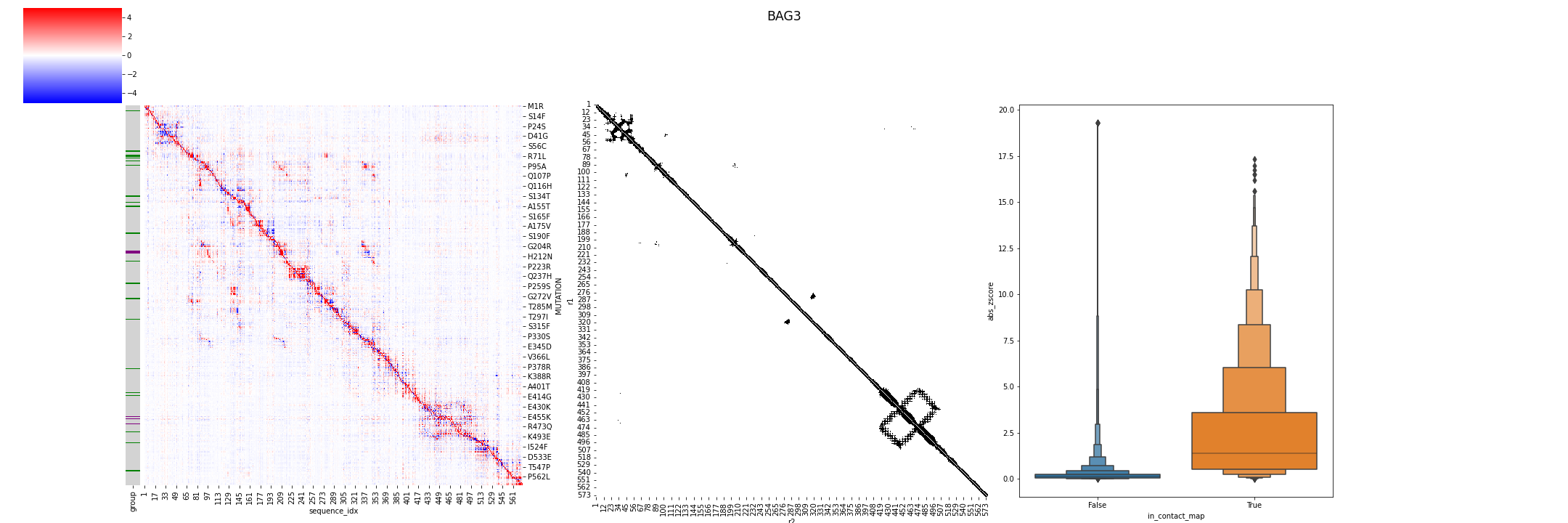}
  \caption{Summary statistics of secondary mutation effects on a variant genetic background segregate in structural features and resemble a contact map. Normalized mean of mutation effects in each position of the BAG Cochaperone 3 (\textit{BAG3}) gene (left) bring out structural patterns consistent with the 10 angstroms contact map of the predicted AlphaFold structure (center; see square pattern corresponding to the BAG domain) and highlight additional, potentially missed structural patterns in regions deemed disordered by AlphaFold. High and low effects tend to segregate within the contact map (right)}
  \label{fig:bag3_contact_map}
\end{figure}

\begin{figure}
  \centering
  \includegraphics[width=135mm]{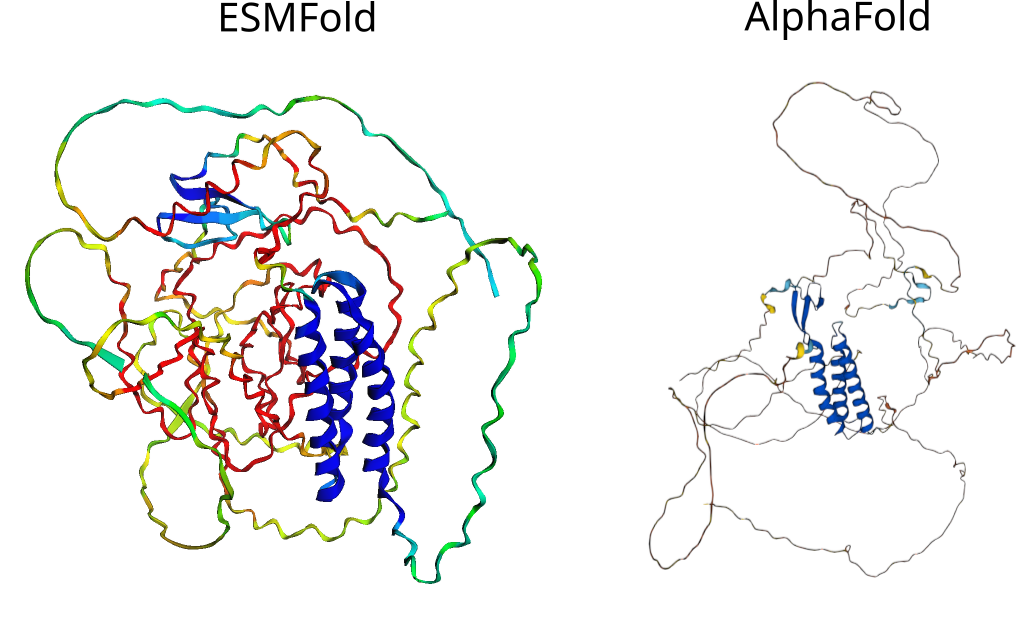}
  \caption{\textit{BAG3} predicted structures via ESMfold (left) and AlphaFold (right). ESMfold picks up more structural features than AlphaFold. Both structures are colored by confidence, from blue (high confidence) to yellow (medium confidence) to red (very low confidence)}
  \label{fig:bag3_contact_map}
\end{figure}

\section{Discussion and Future Directions}
\label{discussion}

Our findings add additional support to the promise of protein language models as tools for interrogating possibly pathogenic genetic variation, following lines of evidence in deep mutational scanning and genome-wide scoring of genetic variants. While extensive large-scale experimental measurements in deep mutational scan data sets provide clear and robust genotype-phenotype maps, curation of clinically relevant genetic variants poses significant challenges. For example, a non-trivial fraction of pathogenic or likely pathogenic ClinVar variants may have incomplete penetrance and therefore their pathogenicity may be highly context dependent. We focused on an expert curated subset of medically actionable genes to enrich our genetic variant dataset for better studied variants with well-established genotype-phenotype associations. Likewise, medical phenotypes in ACMG gene list include diseases with highly complex diagnosis criteria, and it is challenging to establish causal associations with every single genetic variant in these genes. Population scale biomedical databases and biobanks such as UK Biobank (\cite{ukbb}) provide an opportunity to refine the curation the genetic and phenotypic data from hundreds of thousands of individuals.

Additionally, we explore an \textit{in silico} search of compensatory mutations using protein language models as a means to further characterize the effect of genetic variation. This compensatory score maps reveal potentially unappreciated structural features in some cases. For this work, we only considered single rescue mutations, however, it is conceivable that compensatory interactions may involve more than one amino acid substitution within the same protein, co-evolving changes in the interacting protein partners, or more subtle synonymous changes affecting the secondary structure. It remains elusive to what extent incorporating higher order dependencies between sites can help improve the predictive accuracy of protein language models or highlight any potential limitation. Masked language models provide a convenient extended co-masking strategy to probe the relevance of higher order interactions. Rescue experiments offer a powerful framework to diagnose the limitations of large language models to capture clinically relevant aspects of complex fitness landscapes. While deep mutational scan experiments provide an unbiased estimate of the mutational landscape, we encourage the community to generate and curate rescue mutation experiments to inform the design of more sophisticated co-masking strategies and leverage large language models more effectively for downstream clinical prediction tasks.

\section{Data and code availability}
All the data used in this paper are publicly available. Details on datasets, models and analysis code can be found at \url{https://github.com/dimenwarper/llm-for-clinical-variants}. 

\bibliography{neurips_2022}

\appendix

\section{Appendix}

List of ACMG genes considered in this study:

\begin{tabular}{ | c | c | c | c | c | c | c | } 
  \hline
  ACTA2 & ACTC1 & ACVRL1 & BAG3 & BMPR1A & BTD \\
  \hline
  CASQ2 & DES & DSC2 & ENG & GAA & GLA \\
  \hline
  HFE & HNF1A & KCNQ1 & LDLR & LMNA & MAX \\
  \hline
  MEN1 & MLH1 & MSH2 & MUTYH & MYL2 & MYL3 \\ 
  \hline
  NF2 & OTC & PCSK9 & PKP2 & PRKAG2 & PTEN \\
  \hline
  RB1 & RPE65 & SDHAF2 & SDHB & SDHC & SDHD \\
  \hline
  SMAD3 & SMAD4 & STK11 & TGFBR1 & TGFBR2 & TMEM127 \\
  \hline
  TMEM43 & TNNC1 & TNNI3  & TNNT2 & TP53  & TPM1  \\
  \hline
  TRDN   & TTR     & VHL    & WT1   &        &  \\
  \hline
\end{tabular}

\renewcommand{\figurename}{Supplementary Figure}
\setcounter{figure}{0}
\begin{figure}[b]
  \centering
  \includegraphics[width=130mm]{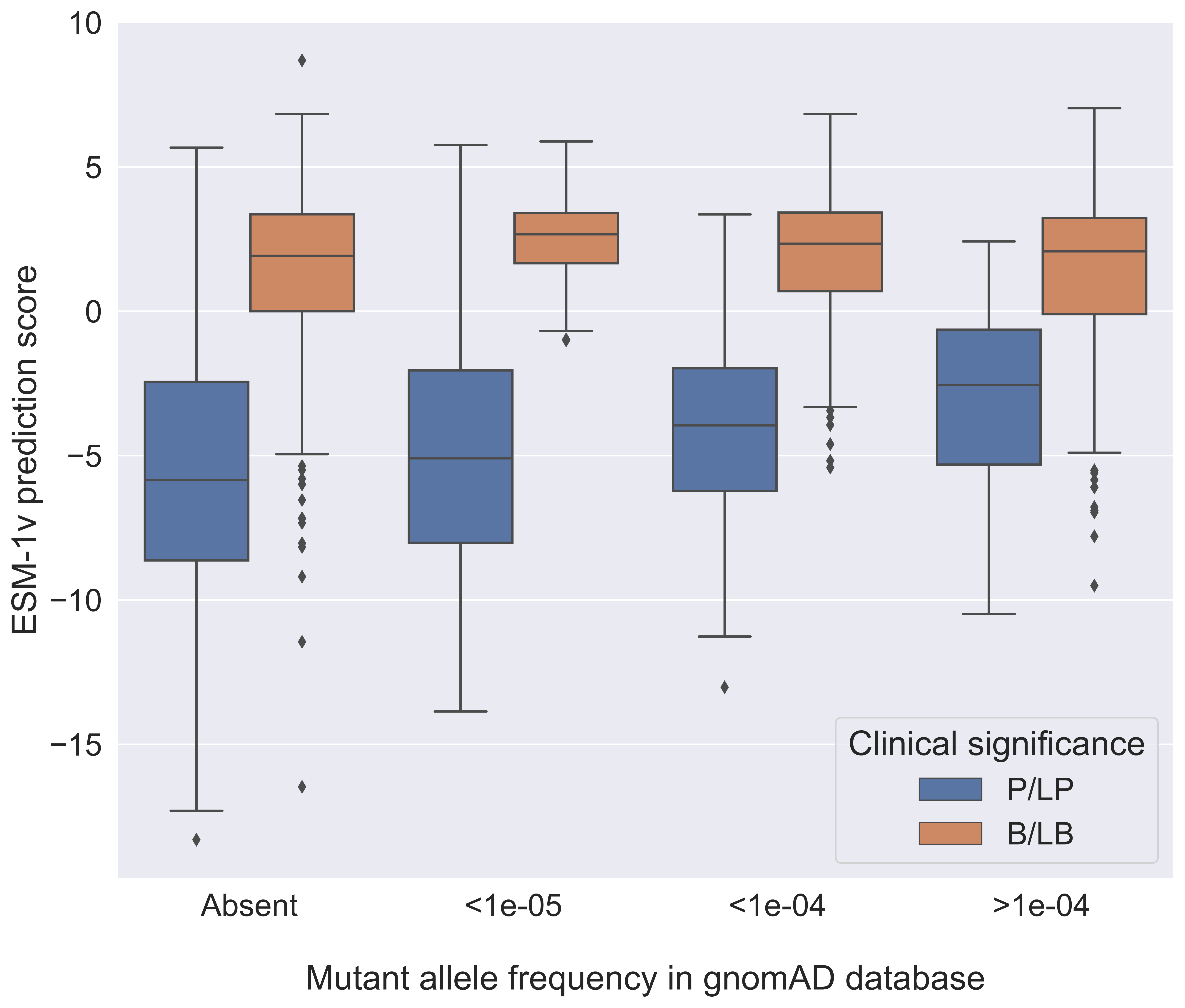}
  \caption{Distribution of ESM-1v prediction scores for pathogenic (P/LP) and benign (B/LB) ClinVar variants group by their respective population allele frequencies in the gnomAD genetic variation database. ESM-1v prediction scores are the average ensemble score of five models. For all pairwise comparisons, two-sided Mann-Whitney U test p-values < 3.5e-05.}
  \label{fig:suppfig1}
\end{figure}

\begin{figure}
  \centering
  \includegraphics[width=130mm]{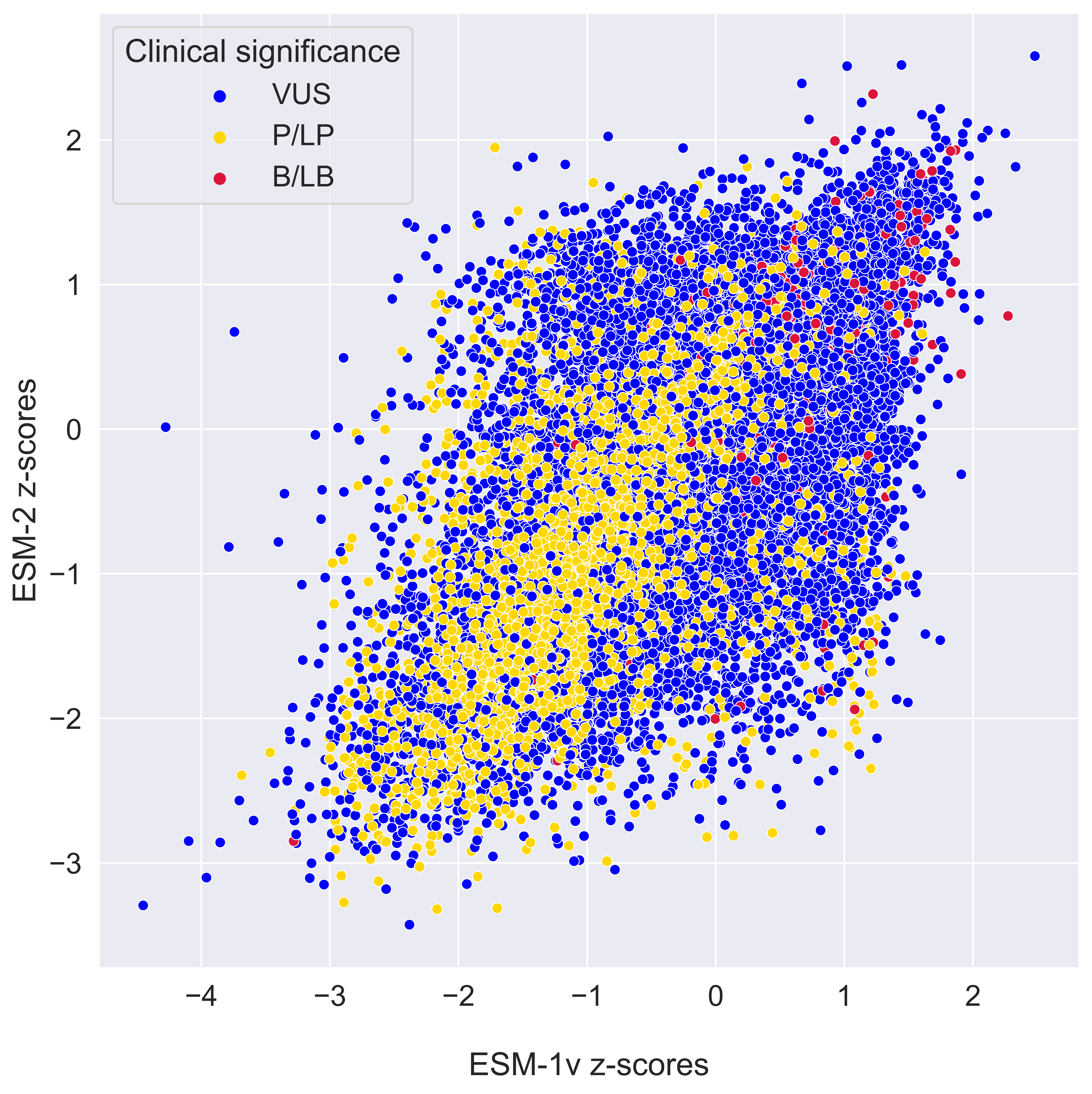}
  \caption{Comparison of ESM-1v and ESM-2 prediction scores as normalized across the entire predictions among ClinVar variants in 53 ACMG genes. Pearson's correlation of $r$=0.66}
  \label{fig:suppfig2}
\end{figure}

\begin{figure}
  \centering
  \includegraphics[width=130mm]{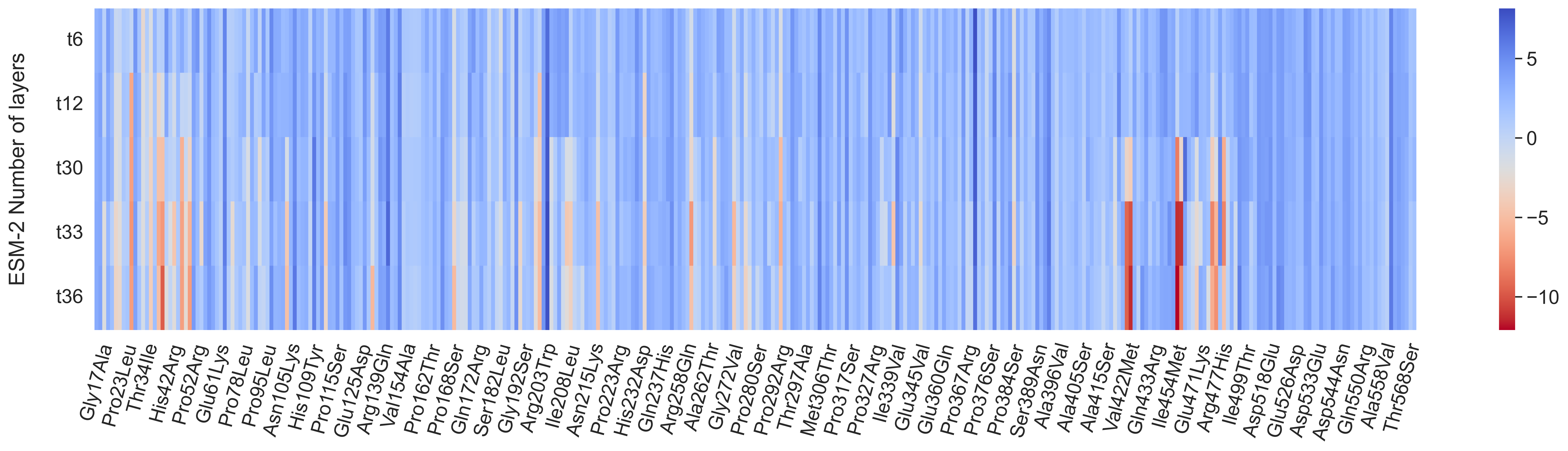}
  \caption{Comparison of ESM-2 prediction scores for \textit{BAG3} variants with unknown significance (VUS) across ESM-2 pre-trained models with varying number of layers.}
  \label{fig:suppfig3}
\end{figure}

\begin{figure}[t]
  \centering
  \includegraphics[width=130mm]{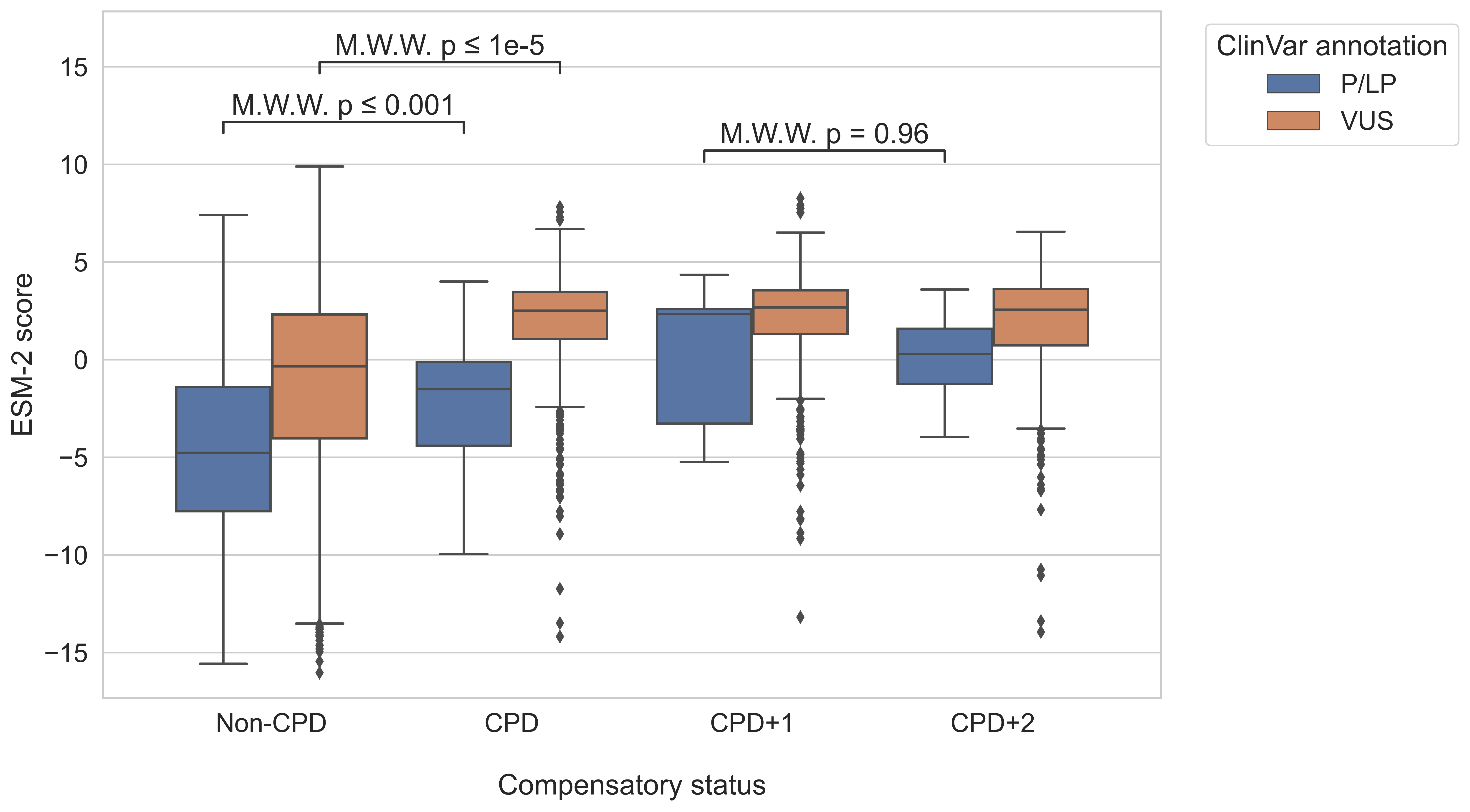}
  \caption{ESM-2 scores for pathogenic variants (P/LP) and variants of unknown significance (VUS) at putative compensated sites. CPD refers to sites where the mutant residue is present in at least one non-human species. CPD+1 and CPD+2 refer to CPDs where the neighboring residue or two residues, respectively, are required to be fully conserved. Mann-Whitney-Wilcoxon (M.W.W) two-sided test p-values are shown.}
  \label{fig:suppfig4}
\end{figure}

\begin{figure}[t]
  \centering
  \includegraphics[width=130mm]{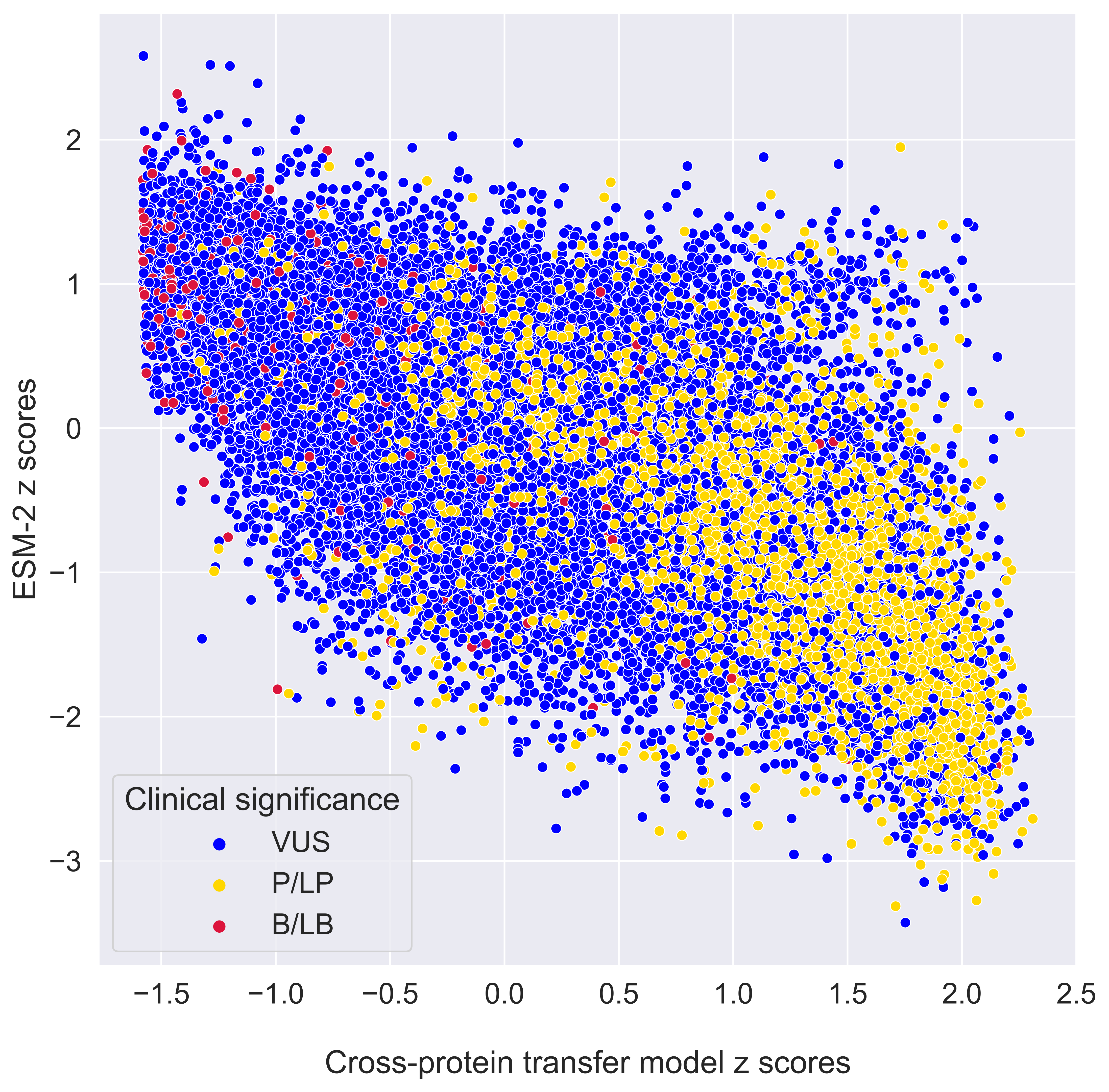}
  \caption{Comparison of ESM-2 and Cross-Protein Transfer (CPT)\cite{cpt} model prediction scores as normalized across the entire predictions among ClinVar variants in 51 ACMG genes for which both models have predictions available. Spearman's rank correlation between the two predictors is $\rho$=0.69.}
  \label{fig:suppfig5}
\end{figure}

\end{document}